\documentclass{article}
\usepackage{ijcai24}

\usepackage{times}
\usepackage{soul}
\usepackage{url}
\usepackage[hidelinks]{hyperref}
\usepackage[utf8]{inputenc}
\usepackage[small]{caption}
\usepackage{graphicx}
\usepackage{amsmath}
\usepackage{amsthm}
\usepackage{booktabs}
\usepackage{algorithm}
\usepackage{algorithmic}
\usepackage[switch]{lineno}
\usepackage{enumitem}
\usepackage{pgfplots}
\usepackage{amssymb}
\usepackage{multirow}
\usepackage{multicol}
\usepackage{autobreak}
\newcommand{\ignore}[1]{}
\usetikzlibrary{patterns}
\title{Enhancing Document-level Translation of Large Language Model \\via Translation Mixed-instructions}

\author{
    Author Name
    \affiliations
    Affiliation
    \emails
    email@example.com
}

\author{
Yachao Li$^1$
\and
Junhui Li$^2$\and
Jing Jiang$^{1}$\And
Min Zhang$^2$\\
\affiliations
$^1$The Key Laboratory of Linguistic and Cultural Computing of Ministry of Education, \\Northwest Minzu University, China\\
$^2$Institute of Artificial Intelligence, \\School of Computer Science and Technology, Soochow University, China \\
\emails
}
\begin{document}

\maketitle

\begin{abstract}

Existing large language models (LLMs) for machine translation are typically fine-tuned on sentence-level translation instructions and achieve satisfactory performance at the sentence level. However, when applied to document-level translation, these models face a significant challenge, particularly when dealing with documents containing over 512 tokens. This challenge arises from the issue of sentence-level coverage, where subsequent sentences in the document remain untranslated. As a result, the document-level translation capability of LLMs fine-tuned on sentence-level translation instructions is significantly limited. We conjecture that the primary cause of LLMs' weak document-level translation performance is the absence of document-to-document mapping ability. To address the issue, we propose an approach that combines sentence-level and document-level translation instructions of varying lengths to fine-tune LLMs. Our proposed translation mixed-instructions enable LLMs (Llama-2~7B and 13B) to maintain consistent translation performance from the sentence level to documents containing as many as 2048 tokens. Extensive experimental results show that the proposed approach significantly enhances the document-level translation capabilities of LLMs on 10 language pairs, effectively mitigating the sentence-level coverage issue in document-level translation. Experimentation on discourse phenomena has demonstrated that our document-level translation approach significantly improves translation quality, both in terms of BLEU score and discourse coherence.

\end{abstract}

\section{Introduction}
\label{sect:intro}

Document-level neural machine translation (NMT) aims to translate the source document into a coherent and fluent target document~\cite{zhang-2018-doc,maruf:2019-doc,voita:2019-doc,zhang-etal-2020-long,bao-etal-2021-g,sun-etal-2022-rethinking,p-transformer}.
Large language models (LLMs), such as ChatGPT~\cite{openai_2022_chatgpt} and GPT-4~\cite{openai_arxiv_2023_gpt4}, demonstrate impressive performance in machine translation, particularly in document-level translation~\cite{moslem-etal-2023-adaptive,wang-etal-2023-document-level,jiao_2023_chatgpttrans,hendy_gpt-mt-2023,zhu2023multilingual}. 
However, these translation experiments are conducted on well-tuned complex LLMs, leaving the question of how lightweight LLMs acquire document-level translation ability unanswered.
Different from previous research, this paper investigates the document-level translation capabilities of lightweight LLMs and presents a simple and effective approach to enhance the document-level translation performance of LLMs.

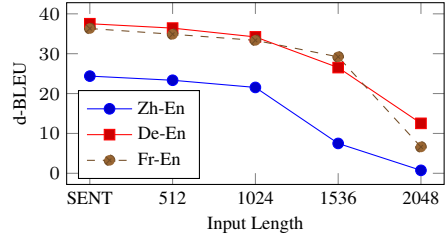
\begin{figure}[!t]
	\setlength{\abovecaptionskip}{0pt}
	\begin{center}
		\pgfplotsset{compat=1.13}
		\begin{tikzpicture}
		\tikzset{every node}=[font=\scriptsize]
		\begin{axis}
		[height=4.0cm,width=6.6cm,enlargelimits=0.07, tick align=inside, legend style={cells={anchor=west},legend pos=south west}, xticklabels={SENT,512,1024,1536,2048},
		xtick={1,2,3,4,5},
		ylabel={ d-BLEU },
		xlabel={Input Length},
		ymax=40,
		ymin=0,
		ytick distance=10,
		]	

		\addplot+[] coordinates
		{(1,24.38) (2,23.34) (3,21.54) (4,7.49) (5,0.74) };
		\addlegendentry{\scriptsize{Zh-En}}

		\addplot+[] coordinates
		{(1,37.52) (2,36.42) (3,34.21) (4,26.52) (5,12.53) };
		\addlegendentry{\scriptsize{De-En}}

		\addplot+[dashed] coordinates
		{(1,36.32) (2,34.88) (3,33.35) (4,29.21) (5,6.61) };
		\addlegendentry{\scriptsize{Fr-En}}
  
		\end{axis}
		\end{tikzpicture}
		\caption{Translation performance of Llama-2 with different input lengths. ``SENT'' denotes the sentence-level input. For a fair comparison, we first obtain the document-level translation of original documents and then compute document-level BLEU (d-BLEU).} 
        \label{fig:probing}
	\end{center}
\end{figure}

To evaluate the document-level translation performance of lightweight LLMs, next we take the widely-used Llama-2~7B model~\cite{llama2} as a representative which is fine-tuned on the Alpaca~\cite{alpaca} instruction set (see Section~\ref{sect:main} for details).
We first segment documents in the test set into sub-documents containing up to $L$ tokens, where $L$ takes values in \{512, 1024, 1536, 2048\}, and then feed the sub-documents into Llama-2 as input. Figure~\ref{fig:probing} shows the translation performance trend between sentence-level and document-level translations for Chinese~/~German~/~French-to-English (Zh~/~De~/~Fr-En). 
Compared to the sentence-level translation, Llama-2 achieves comparable performance when performing document-level translation with $L=512$. However, the translation performance deteriorates significantly when the document length surpasses 512 tokens. For example, the translation almost failed when the input length reached 2048 tokens on Zh-En and De-En tasks.
Additionally, the performance trends for Zh-En, De-En, and Fr-En translations are consistent, indicating that the vanilla Llama-2~7B model struggles with document-level translation.

We conjecture that the reason behind Llama-2's low document-level translation ability lies in the machine translation instructions of Alpaca~\cite{alpaca} which are all at the sentence or phrase level. Below is an example of such instructions:\\
\textit{
\{``instruction": ``Translate the following sentence from English to French.",\\
        ``input": ``I am happy to meet you.",\\
        ``output": ``Je suis heureux de te rencontrer."\}
}\\
Therefore, fine-tuning Llama-2 with the above instructions of Alpaca~\cite{alpaca} may enhance its sentence-level translation ability, but it could hinder its ability to learn effective document-level translation ability.

Motivated by the above experimental results, it is crucial to include the document-level translation instructions during the fine-tuning process of translation-oriented LLMs.
To achieve this, we propose simple yet effective mixed-instructions to improve the document-level translation of LLMs. Our approach combines translation instructions of varying lengths (sentence-level and document-level) during fine-tuning.
By incorporating instructions with different lengths, LLMs can attain robust performance in both sentence-level and document-level translation.
Experimental results on 5 popular document-level translation datasets in 10 language pairs demonstrate that our proposed translation mixed-instructions significantly enhance the document-level translation performance of LLMs.
In summary, the contributions of the paper are three-fold:
\begin{itemize}

\item We identify the weakness in document-level translation of LLMs and propose a simple yet effective approach to enhance their performance by combining sentence-level and document-level translation instructions of varying lengths. Our findings provide valuable insights into translation instruction tuning for LLMs.

\item  Our research reveals that sentence-level and document-level translation instructions play different roles in LLMs, with the former mainly stimulating sentence-level translation ability and the latter enhancing document-level translation ability. Moreover, we find that fine-tuning too many sentence-level translation instructions can severely harm the document-level translation ability of LLMs.

\item  Our extensive experimental results show that the proposed translation mixed-instructions significantly improve the document-level translation performance of LLMs on 10 language pairs, maintaining stable translation performance from sentence-level to inputs of 2048 tokens in length.
    
\end{itemize}

\section{Translation Mixed-instructions Tuning}

Instruction tuning is a process that enhances the capability of a large language model to understand specific tasks. 
This is achieved by training the models using maximum likelihood estimation (MLE) given the task stating ($\texttt{s}$), input ($\texttt{x}$), and output ($\texttt{y}$), as shown in Equ.~\ref{loss}. 
The goal is to enable the model to accurately predict the target output $\texttt{y}=y_t|_{t=1}^{T}$ based on the task stating and input, thereby improving the model's understanding of the translation task.
\begin{equation}
\label{loss}
    L_{mle} = \sum_{t=1}^T Log P(y_t | y_{<t}, \texttt{x}, \texttt{s}).
\end{equation}

The object remains the same as the one used during the training stage.
In the following, we introduce the sentence-level and document-level translation instructions, respectively, and then propose our fine-tuning approach for translation instructions in LLMs.

\begin{figure}[t]
\begin{center}
  \includegraphics[width=7.5cm]{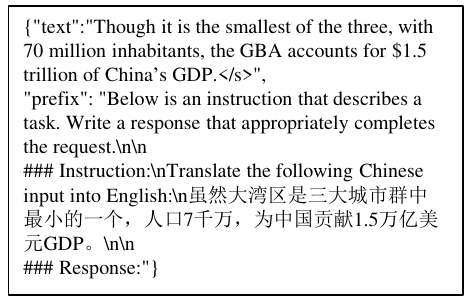}
\end{center}
%\vspace*{3mm}
\caption{Example of sentence-level translation instruction in Jason format.}
\label{fig:sent-inst}
\end{figure}

\begin{figure}[t]
\begin{center}
  \includegraphics[width=7.5cm]{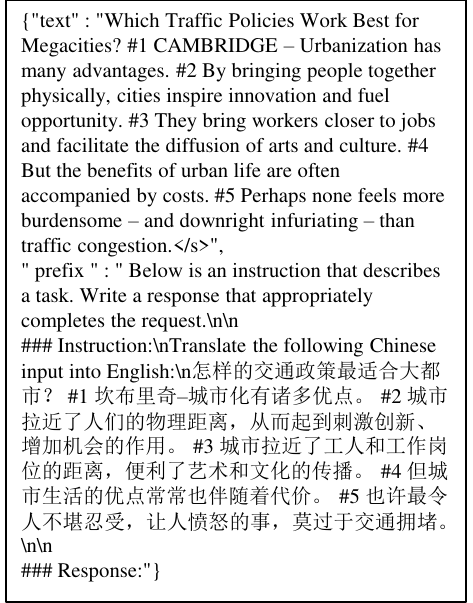}
\end{center}
%\vspace*{3mm}
\caption{Example of document-level translation instruction with a document of 6 sentences.}
\label{fig:doc-inst}
\end{figure}

\subsection{Sentence-level Translation Instruction}

Sentence-level translation instructions are widely used in many general LLMs~\cite{vilar-etal-2023-prompting} and translation-oriented LLMs~\cite{yang-etal-2023-BigTranslate,bayling}. 
As shown in Figure~\ref{fig:sent-inst}, the translation task stating, source language sentence input are concatenated to construct a translation instruction (``\#\#\# Instruction:''), while the target language sentence output serves as the response of an LLM (``text:''), which stimulates LLMs to follow machine translation tasks. Note that we follow the training data format by ~\cite{jiao-etal-2023-parrot}, which is different from the original Alpaca~\cite{alpaca} data format.

Selecting sentence-level input and output for constructing translation instructions is a logical choice, considering that most machine translation data sets exist in bilingual text at the sentence level. 
This approach allows LLMs to learn the mapping between source and target language, a widely adopted approach that has achieved effective translation performance.

\subsection{Document-level Translation Instruction}

As discussed in Section~\ref{sect:intro}, it is very important to include document-level translation instructions during fine-tuning.
As depicted in Figure~\ref{fig:doc-inst}, we form documents by concatenating consecutive sentences and then create document-level translation instructions. In Figure~\ref{fig:doc-inst}, \#1, \#2,~$\cdots$ represent sentence separators, which enable the recovery of sentence-to-sentence alignment between the source and target input (see Section~\ref{sect:ana-cover} for further discussion).
It is important to note that the instruction does not clearly distinguish between sentence-level and document-level instructions. The only difference between the two lies in the length of the source and target language inputs.

\subsection{Sentence- and Document-level Translation Mixed-instructions Tuning}

We conjecture that sentence-level and document-level translation instructions serve distinct functions. To encourage both sentence-level and document-level translations of LLMs, we propose combining both instruction types to fine-tune LLMs.
Specifically, given a document-level bitext, we aim to stimulate the translation ability of LLMs on documents of varying lengths. We achieve this by splitting long documents into sub-documents with a maximum of $L$ tokens, where $L\in\{512, 1024, 1536, 2048\}$. Then, we combine sentence-level and document-level translation instructions of different lengths to create \textit{translation mixed-instructions}.

The maximum document length is contingent upon the longest input length supported by the LLMs. For instance, Llama-2's maximum supported length is 4096, so we set the maximum input document length to 2048. 
In sub-document segmentation, we choose the length segmentation interval of 512, please see Section~\ref{fig:d-inst} for more discussions.
By providing translation instructions of varying lengths, the LLMs can learn to map input and output translations of different lengths. This enables the LLMs to acquire robust translation abilities at both sentence level and document level.

\section{Experimentation}

\subsection{Experimental Setup}
\noindent\textbf{Datasets.} 
To avoid data leakage~\cite{garcia_data_leak}, we utilize the latest News Commentary v18 (News18) from WMT 2023,\footnote{https://www.statmt.org/wmt23/translation-task.html} which features parallel text with document boundaries.
We conduct experiments on multiple document-level translation tasks across five language pairs in both directions: Chinese-English (Zh-En), German-English (De-En), French-English (Fr-En), Spanish-English (Es-En), and Russian-English (Ru-En). 
For each language pair, we randomly shuffle the corpus documents, selecting 80\% as the training set. For the development set, we select 10\% of the corpus and then select 150 documents from it as the development set. The test set is constructed in the same way as the development set.
The statistics of the datasets are outlined in Table~\ref{tbl:datasets}.\footnote{We will release the dataset split for re-implementation.}
Following~\cite{jiao-etal-2023-parrot}, we process the bi-lingual data into machine translation instructions format for LLMs tuning, as shown in Figure\ref{fig:sent-inst}.
In the pre-processing stage for conventional NMT models, we tokenize all sentences into sub-words using a joint SentencePiece with 30K operations~\cite{sentencepiece}. However, for experiments involving Llama-2, we utilize the tokenizer provided by the model itself.

\begin{table}[!t]
\begin{center}
\begin{tabular}{l|cc}
    \hline
\multirow{2}* {Dataset} &\multirow{2}* \bf \#DOC & \multirow{2}* \bf \#SENT \\
& \small{ (train / valid / test) }& \small{ (train / valid / test) } \\
\hline
    Zh-En & \small{ 8.6K / 150 / 150 } & \small{ 342K / 6.0K / 5.9K } \\ \hline
    De-En & \small{ 8.4K / 150 / 150} & \small{ 333K / 5.8K / 5.9K } \\ \hline
    Fr-En & \small{ 7.9K / 150 / 150} & \small{ 310K / 5.9K / 5.8K } \\ \hline
    Es-En & \small{ 9.6K / 150 / 150 } & \small{ 378K / 5.8K / 5.8K } \\ \hline
    Ru-En & \small{ 7.3K / 150 / 150 } & \small{ 272K / 5.7K / 5.6K } \\ \hline
\end{tabular}
\end{center}
\caption{\label{tbl:datasets} Statistics on the document-level translation datasets.} 
\end{table}

\begin{table*}[!ht]
%\small
\begin{center}
  \begin{tabular}{lll|cc|cc|cc|cc|cc}
    \hline \multirow{2}{*}{\# } & \bf \multirow{2}{*}{System} & \bf \multirow{2}{*}{Input} & \multicolumn{2}{c}{\bf Zh-En} & \multicolumn{2}{c}{\bf De-En}  &  \multicolumn{2}{c}{\bf Fr-En } & \multicolumn{2}{c} { \bf Es-En} & \multicolumn{2}{c}{ \bf Ru-En} \\ 
     
     & & & \scriptsize{BLEU} & \scriptsize{COMET} & \scriptsize{BLEU} & \scriptsize{COMET} & \scriptsize{BLEU} & \scriptsize{COMET} & \scriptsize{BLEU} &\scriptsize{ COMET}&\scriptsize{BLEU}&\scriptsize{COMET} \\ \hline \hline 
    \multicolumn{13}{c}{\textbf{Sentence-level Translation} (s-BLEU)} \\ \hline
   1&{P-Trans} & \small SENT &  25.72 & 0.8419 & 38.53 & 0.8732 & 35.26 & 0.8595 & 45.43 & 0.8805 & 29.75 & 0.8232 \\
   \hline
   \multicolumn{13}{c}{Llama-2 7B} \\
   2&{Alpaca} & SENT & 22.63 & 0.8521 & 36.43 & 0.8835 &  34.85 & 0.8745 & 42.27 & 0.8885 & 29.63 & 0.8454 \\
   \hline
   3&{S-INST} & SENT & 27.43 & 0.8633 & 40.93 & 0.8913 & 38.18 & 0.8804 & 47.52 & 0.8969 & 33.00 & 0.8504 \\
   4&{M-INST} & SENT &  
 27.71 & 0.8639 & 41.19 & 0.8913 & 38.20 & 0.8803 & 47.88 & 0.8971 & 33.09 & 0.8509 \\
    \hline
    \multicolumn{13}{c}{Llama-2 13B} \\
   5&{S-INST} & SENT & 29.62 
 & 0.8684 & \textbf{42.65} & \textbf{0.8946} & \textbf{39.68} & \textbf{0.8839} & 49.58 & \textbf{0.8995} & 34.80 & 0.8545 \\
  6&{M-INST} & SENT & \textbf{29.82} 
 & \textbf{0.8686} & 42.45 & 0.8935 & 39.57 & 0.8826 & \textbf{49.61} & 0.8992 & \textbf{35.14} & \textbf{0.8552} \\
 %--------------------------------
    \hline \hline
    \multicolumn{13}{c}{\textbf{Document-level Translation} (d-BLEU)} \\
    \hline
   7 & \multirow{4}{*}{P-Trans} & 512 & 26.80 & 0.8498 & 39.24 & 0.8737 & 36.39 & 0.8595 &  46.51 & 0.8774 & 34.23 & 0.8434  \\     
   8 & & 1024 & 27.16 & 0.8439 & 39.78 &  0.8724 & 36.72 & 0.8668 & 46.81 & 0.8770 & 34.80 & 0.8459 \\
   9 & & 1536 & \textbf{27.40} & 0.8601 & 40.06 & 0.8744 & 37.01 & 0.8596 & \textbf{47.19} & 0.8732 & 35.18 & 0.8478 \\   
   10 & & 2048 & 27.28 & \textbf{0.8632} & \textbf{40.09} & \textbf{0.8802} & \textbf{37.03} & \textbf{0.8704} & 46.89 & \textbf{0.8809} & \textbf{35.35} & \textbf{0.8542} \\   
    \hline
    \multicolumn{13}{c}{Llama-2 7B} \\

   11 & S-INST & 512 & 13.58 & 0.7658 & 23.11 & 0.7788 & 20.35 & 0.7485 & 30.47 & 0.7811 & 0.13 & 0.5628 \\
    \hline
    
   12 & \multirow{4}{*}{M-INST} & 512 & 28.66 & 0.8545 & \textbf{42.06} & 0.8771 & 39.34 & 0.8692 &  \textbf{48.36} & 0.8778 & 37.44 & 0.8499 \\
   13 & & 1024 & \textbf{28.85} & 0.8828 & 41.95 & \textbf{0.8977} & \textbf{39.44} & \textbf{0.8897} & 48.29 & 0.8975 & \textbf{38.14} & 0.8765 \\ 
   14 & & 1536 & 28.64 & \textbf{0.8875} & 41.98 & 0.8923 & 39.42 & 0.8854 & 48.01 & 0.8908 & 38.10 & \textbf{0.8793} \\   
   15 & & 2048 & 28.41 & 0.8789 & 41.48 & 0.8945 & 38.66 & 0.8818 & 47.54 & \textbf{0.8938} & 37.47 & 0.8716 \\   
    \hline
    \multicolumn{13}{c}{Llama-2 13B} \\

   16 &S-INST & 512 & 0.49 & 0.6040 & 0.10 & 0.5583 & 0.01 & 0.5203 & 0.02 & 0.5315 & 0.02 & 0.4324 \\
     \hline
  17 & \multirow{4}{*}{M-INST} & \small 512 & 31.60 
 & 0.8582 & \textbf{43.95} & 0.8789 & 40.83 & 0.8708 & 50.22 & 0.8796 & 39.71 & 0.8530 \\
   18 & & 1024 & \textbf{31.78} & 0.8839 & 43.90 & \textbf{0.8985} & 41.24 & \textbf{0.8905} & \textbf{50.34} & 0.8992 & 40.27 & 0.8806 \\
   19 & & 1536 & 31.65 & \textbf{0.8908} & 43.90 & 0.8947 & \textbf{41.39} & 0.8878 & 50.25 & 0.8930 & \textbf{40.51} & \textbf{0.8825} \\
   20 & & 2048 & 31.50 & 0.8826 & 43.58 & 0.8971 & 41.07 & 0.8841 & 50.07 & \textbf{0.8959} & 39.88 & 0.8761 \\ 
    
    \hline    
  \end{tabular}
\end{center}
    \caption{\label{tbl:main} Translation performance of the Llama-2 and P-Transformer~(P-Trans) on \textit{X}-En translations}
\end{table*}

\noindent\textbf{Model Settings.} 
We select the popular Llama-2~7B and 13B models as the foundation LLMs~\footnote{https://huggingface.co/meta-llama/Llama-2-7b-hf}. For LLMs training, we utilize LoRA~\cite{hu2022lora} to fine-tune all models by setting lora\_r to 8 and lora\_alpha to 16. Additionally, we apply LoRA target modules to both the query and value components.
We opt for the transformers and DeepSpeed framework for LLMs training. 
We implement conventional NMT models using \textit{Fairseq}~\cite{ott_etal:2018}, and choose the state-of-the-art P-Transformer model~\cite{p-transformer} as our sentence-level and document-level NMT baselines.

\noindent\textbf{Training.} 
For LLMs, the document-level models were trained on a single A100 GPU, while the sentence-level models were trained on two RTX 3090 GPUs. To ensure a fair and consistent evaluation, we fine-tuned all models for one epoch with a batch size of 32 and used the final checkpoint to evaluate the performance. 
To avoid over-fitting in P-Transformer models~\cite{p-transformer}, we employed an early stopping strategy with patience of 5 on the validation set or stopped training after a maximum of 30 epochs. Following P-Transformer~\cite{p-transformer}, we also combined both sentence- and document-level data to train NMT models.

\noindent\textbf{Evaluation.} Following previous related work~\cite{p-transformer}, we report case-sensitive sentence-level BLEU (s-BLEU) for sentence-level translation. We also report the document-level BLEU~\cite{liu-2020-mbart} (d-BLEU) for the document-level translation, which is computed by matching n-grams in the whole document after removing the special tokens (\#1, \#2, $\cdots$). To better show the translation quality, we also adopt the widely-used COMET~\cite{rei-etal-2020-comet} score with \textit{Unbable/wmt22-comet-da},\footnote{https://github.com/Unbabel/COMET} which relies on cross-lingual pre-trained models.

\subsection{Main Results on \textit{X}-En Translations}
\label{sect:main}
Table~\ref{tbl:main} shows the results of \textit{X}-En translations tasks. ``S-INST'' and ``M-INST'' represent the Llama-2 models fine-tuned on sentence-level instructions and mixed-instructions, respectively. ``SENT'' indicates the sentence-level input, and $L\in\{512, 1024, 1536, 2048\}$ denote the document-level input with up to $L$ tokens.
To enhance the translation performance of the vanilla Llama-2 model fine-tuned on Alpaca~\cite{alpaca}, we added an extra 10K sentence-level translation instructions to the  Alpaca~\cite{alpaca} set, which is denoted as ``Alpaca'' in the table. Note that the translation results in Figure~\ref{fig:probing} used this data set.

\noindent\textbf{Results on Sentence-Level Translation.}
The sentence-level translation results in Table~\ref{tbl:main} demonstrate that the conventional NMT baseline (\#1) has achieved good translation performance. In addition, the vanilla Llama-2~7B baseline (\#2) is lower than that of the NMT baseline (\#1) in BLEU, but higher in COMET.

The Llama-2~7B model shows comparable sentence-level translation performance when fine-tuned on sentence-level translation instructions (\#3) or document-level translation instructions (\#4). Notably, both versions of the Llama-2~7B model significantly outperformed the P-Transformer (\#1) and LLM baseline (\#2).
The experimental results indicate that sentence-level translation instructions can effectively enhance the sentence-level translation performance of the Llama-2~7B model. Additionally, the results show that including document-level translation instructions does not further improve sentence-level translation performance.

We also conducted experiments on a larger Llama-2~13B model. As shown in Table~\ref{tbl:main}~(\#5 and \#6), the translation performance is significantly improved across 5 language pairs, demonstrating the larger LLM's ability to enhance translation performance.
It is worth noting that the translation performances are similar between Llama-2~13B fine-tuned on sentence-level translation instructions~(\#5) and document-level translation instructions~(\#6). These results further support the idea that sentence-level and document-level translation instructions serve distinct functions.

\begin{table*}[!ht]
%\small
\begin{center}
  \begin{tabular}{lll|cc|cc|cc|cc|cc}
    \hline \multirow{2}{*}{\# } & \bf \multirow{2}{*}{System} & \bf \multirow{2}{*}{Input} & \multicolumn{2}{c}{\bf En-Zh} & \multicolumn{2}{c}{\bf En-De}  &  \multicolumn{2}{c}{\bf En-Fr } & \multicolumn{2}{c} { \bf En-Es} & \multicolumn{2}{c}{ \bf En-Ru } \\ 
     
     & & & \scriptsize{BLEU} & \scriptsize{COMET} & \scriptsize{BLEU} & \scriptsize{COMET} & \scriptsize{BLEU} & \scriptsize{COMET} & \scriptsize{BLEU} &\scriptsize{ COMET}&\scriptsize{BLEU}&\scriptsize{COMET} \\ \hline \hline 
    \multicolumn{13}{c}{\textbf{Sentence-level Translation} (s-BLEU)} \\ \hline
   1&{P-Trans} & \small SENT & \textbf{41.19} & 0.8283 & \textbf{28.55} & 0.7507 & \textbf{31.49} & 0.7836 & \textbf{41.47} & 0.8028 & \textbf{21.53} & 0.8159 \\
   \hline
   \multicolumn{13}{c}{Llama-2 7B} \\
   2&{S-INST} & SENT & 35.60 & 0.8545 & 23.80 & 0.8691 & 29.07 & 0.8630 & 39.38 & 0.8827 & 17.65 & 0.8650 \\
   3&{M-INST} & SENT & 36.41 
  & \textbf{0.8566} & 24.36 & \textbf{0.8707} & 29.31 & \textbf{0.8632} & 39.64 & \textbf{0.8836} & 18.08 & \textbf{0.8673} \\
 %--------------------------------
    \hline \hline
    \multicolumn{13}{c}{\textbf{Document-level Translation} (d-BLEU)} \\
    \hline
   4 & \multirow{4}{*}{P-Trans} & 512 & 50.27 
  & 0.8532 & \textbf{26.14} & 0.8627 & 26.70 & 0.8565 & 38.56 & 0.8568 & 21.46 & 0.9086 \\ 
   5 & & 1024 & 51.69 
  & 0.8615 & 25.94 & 0.8647 & 27.13 & 0.8701 & 39.33 & 0.8590 & 21.73 & 0.9152 \\
   6 & & 1536 & 52.35 
  & 0.8690 & 25.87 & 0.8597 & 27.09 & 0.8466 & 39.20 & 0.8494 & 21.18 & 0.9055 \\   
   7 & & 2048 & \textbf{52.40}  
  & \textbf{0.8720} & 26.07 & \textbf{0.8764} & \textbf{27.39} & \textbf{0.8754} & \textbf{39.93} & \textbf{0.8644} & \textbf{21.76} & \textbf{0.9218} \\   
    \hline
    \multicolumn{13}{c}{Llama-2 7B} \\
    \hline

   8 & S-INST & 512 &  
 0.43 & 0.5898 & 2.49 & 0.6236 & 4.83 & 0.6304 & 0.70 & 0.5709 & 0.04 & 0.6001 \\
    \hline
    
   9 & \multirow{4}{*}{M-INST} & 512 & 36.11 
  & 0.8358 & 23.74 & 0.8684 & 29.68 & 0.8687 & 39.50 &  0.8648 & \textbf{19.53} & 0.8659 \\
   10 & & 1024 & \textbf{37.16} 
  & 0.8578 & \textbf{23.83} & \textbf{0.8819} & \textbf{29.85} & \textbf{0.8727} & \textbf{39.53} & \textbf{0.8794} & 19.40 & \textbf{0.8862} \\ 
   11 & & 1536 & 37.13 
  & \textbf{0.8693} & 23.55 & 0.8764 & 29.51 & 0.8713 & 39.21 & 0.8729  & 19.35 & 0.8837 \\   
   12 & & 2048 & 36.78  
  & 0.8572 & 23.17 & 0.8765 & 29.15 & 0.8647 & 38.42 & 0.8709 & 19.03 & 0.8801 \\   
    \hline    
  \end{tabular}
\end{center}
    \caption{\label{tbl:main2} Translation performance of the Llama-2 and P-Transformer (P-Trans) on En-X translations.}
\end{table*}

\noindent\textbf{Document-level Translation of P-Transformer and Sentence-Level Instruction.}
We divide the test set into different lengths, ranging from 512, 1024, 1536 to 2048 tokens, and compare their translation performance. 
Please note that the maximum sequence length of Llama-2 is 4096, thus the maximum input document length of our paper is set to 2048.
The document-level translation results (\#7 $\sim$ \#10) in Table~\ref{tbl:main} show that the P-Transformer~(P-Trans) baseline has achieved good translation performance across all input lengths. Notably, when the document length reaches 2048 tokens, it still maintains stable translation performance. Generally, a long input document tends to result in higher BLEU and COMET scores.

The performance of Llama-2~7B fine-tuned on sentence-level translation instructions~(\#11) is very low on documents with a length of 512 tokens, which is significantly lower than that of P-Transformer.
Moreover, for the larger Llama-2~13B model (\#16), the translation almost fails on 512 tokens input, and the BLEU is extremely low.
Translation analysis shows that the translation of the first one or few sentences in the document is good, but the subsequent sentences can not be translated effectively, which means that these sentences remain untranslated. Following previous related work~\cite{p-transformer}, we say that these is \textit{sentence-level coverage} issue, as discussed in Section~\ref{sect:ana-cover}.
Experimental results show that sentence-level translation instructions can not effectively stimulate the document-level translation ability of LLMs.

\noindent\textbf{Document-level Translation of Translation Mixed-instructions.}
Table~\ref{tbl:main} (\#12 $\sim$ \#15) presents the document-level translation results of LLama-2~7B fine-tuned on mixed-instructions.
The experimental results show that the model achieves good translation performance on documents ranging from 512 to 2048 tokens. Moreover, the document-level translation performance of Llama-2~7B outperforms that of the P-Transformer baseline over 5 language pairs. Experimental results show that with the help of mixed-instructions instructions, the Llama-2~7B model shows strong document-level translation ability.

Fine-tuning larger Llama-2~13B on mixed-instructions (\#17 $\sim$ \#20), the document-level translation performance significantly improves compared to that of Llama-2~7B over 5 language pairs. Experimental results further confirm that LLMs relay on mixed-instructions to stimulate the document-level translation ability.

\subsection{Main Results on En-\textit{X} Translations}

Table~\ref{tbl:main2} shows the results of En-\textit{X} translation tasks. On the whole, the translation quality in English-X directions is inferior to that in X-English directions, which is in line with the previous findings~\cite{zhang_prompt_2023,zhu2023multilingual}. Please be aware that during the English-Chinese translation evaluation, we translate the Chinese translation into characters first, and then determine the BLEU score.

In sentence-level translation tasks, the results of LLMs (\#2 and \#3) are lower than those of P-Transformer (\#1) in BLEU score, but the former is higher than the latter in COMET score. Consistently, the Llama-2~7B model shows comparable sentence-level translation performance when fine-tuned on sentence-level translation instructions (\#2) or document-level translation instructions (\#3).

In document-level translation tasks, the P-Transformer (P-Trans) demonstrates consistent translation performance on documents of varying lengths (\#4 $\sim$ \#7). Notably, Llama-2~7B, fine-tuned on sentence-level translation instructions (\#8), exhibits poor performance on documents with a length of 512 tokens, significantly inferior to P-Transformer's performance. 
In contrast, Llama-2~7B, fine-tuned on mixed-instructions (\#9 $\sim$ \#12), displays stable translation performance across documents ranging from 512 to 2048 tokens. 
When compared to conventional document-level NMT (\#4 $\sim$ \#7), LLMs yield lower BLEU and COMET scores for En-Zh and En-Ru translations. For En-De and En-Es translations, the BLEU scores of Llama-2~7B are inferior to P-Transformer, but their COMET scores are higher.
However, in English-French translation, the BLEU score of Llama-2~7B is significantly higher, while its COMET score is slightly lower.

\section{Analysis}

In this section, we take Llama-2~7B model as a representative to discuss the effectiveness of the proposed translation mixed-instructions.

\subsection{Analysis on Discourse Phenomena}

To assess whether document-level translation of LLMs genuinely acquires useful contextual information to enhance discourse coherence, we utilize the Zh-En linguistic feature test set provided by~\cite{sun-etal-2022-rethinking} to evaluate various discourse phenomena, including tense consistency (TC), conjunction presence (CP), and pronoun translation (PT). TCP, an overall score calculated as the geometric mean of TC, CP, and PT, has a strong correlation with human evaluation, as demonstrated by~\cite{sun-etal-2022-rethinking}. All results are based on the Llama-2~7B model fine-tuned on the translation mixed-instructions.

\begin{table}[t]

\begin{center}
\small
\begin{tabular}{l|cc|ccc|c}
\hline 
\bf Input & \bf d-BLEU& \bf COMET & \bf TC & \bf CP & \bf PT  & \bf TCP \\ 
\hline
SENT & 21.62 & 0.8479 & 46.5 & 33.8 & 63.5 & 46.4 \\
\hline
512 & 24.89 & 0.8579 & 51.2 & 35.8 & \textbf{67.9} & 49.9 \\ 
1024 & \textbf{24.98} & 0.8555 & 52.1 & \textbf{36.5} & 64.5 & 49.7 \\
1536 & 24.65 & \textbf{0.8610} & \textbf{52.3} & 36.3 & 65.9 & \textbf{50.0} \\
2048 & 24.55 & 0.8604 & 49.1 & 35.6 & 64.9 & 48.4 \\
\hline
\end{tabular}
\end{center}
\caption{\label{tbl:dis-zhen} Discourse phenomena evaluation on the Zh-En test set. TC, CP, PT, and TCP are indicated by accuracy~(\%). We format the translations into the original document and then calculate d-BLEU. }
\end{table}

As shown in Table~\ref{tbl:dis-zhen}, the document-level translations (512 to 2048) improve not only the translation performance in BLEU and COMET but also the discourse phenomena performance, compared to sentence-level translation (SENT). However, when the input document reaches 2048 tokens in length, the discourse phenomena performance declines.

\subsection{Analysis on Sentence-level Instructions}

To analyze the impact of sentence-level translation instructions on the translation capability of LLM, we extract 1K and 3K documents from the training set and process them into sentence-level instructions. 
We then compare the performance differences between document-level translation under two settings.

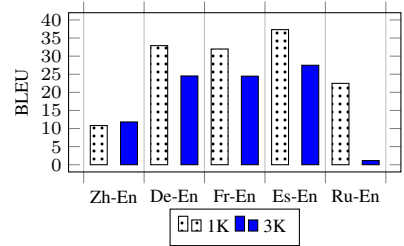
\begin{figure}[!t]
\setlength{\abovecaptionskip}{0pt}
\begin{center}
\pgfplotsset{height=3.7cm,width=6.0cm,compat=1.13}
\begin{tikzpicture}
\tikzset{every node}=[font=\scriptsize]
\begin{axis}[
	ylabel=BLEU,
	enlargelimits=0.05,
	tick align=inside,
	legend style={at={(0.5,-0.23)},
	anchor=north,legend columns=-1},
	ybar interval=0.56,
	xticklabels={Zh-En,De-En,Fr-En,Es-En,Ru-En},
	xtick={1,2,3,4,5,6},
	ytick distance=5,
 	ymax=40,
	ymin=0,
]
\addplot+[color=black,fill=black,pattern=dots,pattern color=black] coordinates{(1,10.82) (2,32.90) (3,31.97) (4,37.33) (5,22.48) (6,0)};
\addplot+[color=black,fill=blue] coordinates{(1,11.81) (2,24.52) (3,24.47) (4,27.48) (5,1.14) (6,0)};
\legend{1K,3K}
\end{axis}
\end{tikzpicture}
\caption{Document-level (512-length tokens) translation performance of Llama-2 fine-tuned on sentence-level instructions.}
\label{fig:ana-doc}
\end{center}
\end{figure}

Figure~\ref{fig:ana-doc} illustrates the performance of document-level translation for 512-length tokens. Our analysis reveals that the translation performance improvements coincide with an increase in translation instructions on the Zh-En task.
Notably, the Llama-2 training data contains limited Chinese data sets. Therefore, the increase in Chinese instructions enhances Zh-En translation.
Conversely, when we increase sentence-level translation instructions for De-En, Fr-En, Es-En, and Ru-En translations, document-level translation performance deteriorates significantly. This suggests that LLM becomes overly reliant on sentence-to-sentence mapping due to excessive sentence-level translation instructions, ultimately impairing its document-to-document translation capability.

\label{sect:ana-cover}

\subsection{Analysis on Document-level Instructions}
\label{fig:d-inst}

To delve deeper into the role of document-level translation instructions, we extract 1K and 3K documents from the training set and process them into 512-length tokens instruction. Subsequently, we fine-tune the Llama-2~7B model on these document-level instructions. Finally, we compare the performance of sentence-level translation and document-level translation under two distinct settings.

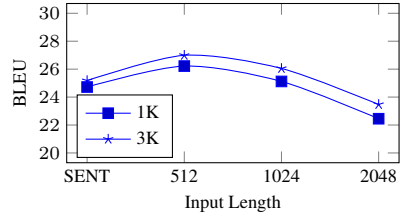
\begin{figure}[t]
	\setlength{\abovecaptionskip}{0pt}
	\begin{center}
		\pgfplotsset{compat=1.13}
		\begin{tikzpicture}
		\tikzset{every node}=[font=\scriptsize]
		\begin{axis}[height=3.7cm,width=6.0cm,enlargelimits=0.07, tick align=inside, legend style={cells={anchor=west},legend pos=south west}, xticklabels={SENT,512,1024,2048},
		xtick={1,2,3,4},
		ylabel={ BLEU },
		xlabel={Input Length},
		ymax=30,
		ymin=20,
		ytick distance=2,
		]	
		\addplot+[smooth,color=blue,mark color=blue,mark=square*] coordinates
		{(1,24.72) (2,26.22) (3,25.12) (4,22.45) };
		\addlegendentry{\scriptsize{1K}}
		\addplot+[smooth,color=blue,mark color=blue,mark=star] coordinates
		{(1,25.18) (2,26.99) (3,26.05) (4,23.47) };
		\addlegendentry{\scriptsize{3K}}
		\end{axis}
		\end{tikzpicture}
		\caption{ The Zh-En translation. } 
        \label{fig:ana-doc-zhen}
	\end{center}
\end{figure}

\begin{figure}[t]
	\setlength{\abovecaptionskip}{0pt}
	\begin{center}
		\pgfplotsset{compat=1.13}
		\begin{tikzpicture}
		\tikzset{every node}=[font=\scriptsize]
		\begin{axis}[height=3.7cm,width=6.0cm,enlargelimits=0.07, tick align=inside, legend style={cells={anchor=west},legend pos=south west}, xticklabels={SENT,512,1024,2048},
		xtick={1,2,3,4},
		ylabel={ BLEU },
		xlabel={Input Length},
		ymax=40,
		ymin=30,
		ytick distance=2,
		]	
		\addplot+[smooth,color=blue,mark color=blue,mark=square*] coordinates
		{(1,36.05) (2,37.45) (3,36.96) (4,35.08) };
		\addlegendentry{\scriptsize{1K}}
		\addplot+[smooth,color=blue,mark color=blue,mark=star] coordinates
		{(1,36.58) (2,38.08) (3,37.66) (4,35.57) };
		\addlegendentry{\scriptsize{3K}}
		\end{axis}
		\end{tikzpicture}
		\caption{ The Fr-En translation. } 
        \label{fig:ana-doc-fren}
	\end{center}
\end{figure}
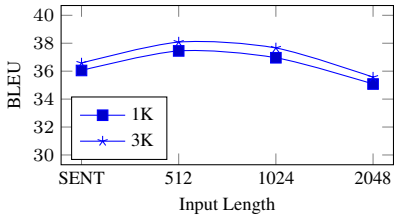

As depicted in Figures~\ref{fig:ana-doc-zhen} and ~\ref{fig:ana-doc-fren}, the translation performances of documents with lengths of 512, 1024, 2048, and sentence level exhibit improvements when the instructions are increased from 1K to 3K on two tasks.
Notably, only using the 512-length document-level translation instructions benefits translation performance even for 2048 tokens input.
The experimental results clearly indicate that the 512-length document-level translation instructions can enhance both sentence-level translation and document-level translation.
Moreover, an interesting finding is that when using 512-length instructions, the 512-length input obtains the best translation performance, which indicates that different lengths of translation instruction can promote the translation performance of the corresponding length. So, we set a document length segmentation interval of 512 to balance the data size and translation performance, achieving good results.

\ignore{
\begin{figure}[ht]
	\setlength{\abovecaptionskip}{0pt}
	\begin{center}
		\pgfplotsset{compat=1.13}
		\begin{tikzpicture}
		\tikzset{every node}=[font=\scriptsize]
		\begin{axis}[height=4.3cm,width=7.0cm,enlargelimits=0.07, tick align=inside, legend style={cells={anchor=west},legend pos=north east}, xticklabels={SENT,512,1024,2048},
		xtick={1,2,3,4},
		ylabel={ BLEU },
		xlabel={Input Length},
		ymax=45,
		ymin=30,
		ytick distance=2,
		]	
		\addplot+[smooth,color=blue,mark color=blue,mark=square*] coordinates
		{(1,38.31) (2,39.60) (3,38.72) (4,36.43) };
		\addlegendentry{\scriptsize{1K}}
		\addplot+[smooth,color=blue,mark color=blue,mark=star] coordinates
		{(1,38.89) (2,40.38) (3,39.64) (4,35.58) };
		\addlegendentry{\scriptsize{3K}}
		\end{axis}
		\end{tikzpicture}
		\caption{ The De-En translation. } 
        \label{fig:ana-doc-deen}
	\end{center}
\end{figure}
}

\subsection{The Sentence-level Coverage Issue}
\label{sect:ana-cover}
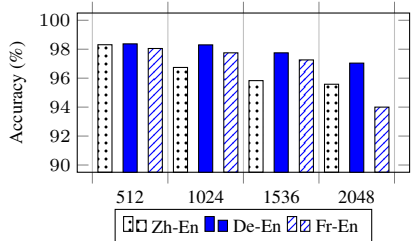
\begin{figure}[!t]
\setlength{\abovecaptionskip}{0pt}
\begin{center}
\pgfplotsset{height=3.7cm,width=6.0cm,compat=1.13}
\begin{tikzpicture}
\tikzset{every node}=[font=\scriptsize]
\begin{axis}[
	ylabel=Accuracy~(\%),
	enlargelimits=0.05,
	tick align=inside,
	legend style={at={(0.5,-0.23)},
	anchor=north,legend columns=-1},
	ybar interval=0.56,
	xticklabels={512,1024,1536,2048,8},
	xtick={1,2,3,4,5},
	ytick distance=2,
 	ymax=100,
	ymin=90,
]
% Zh-En
\addplot+[color=black,fill=black,pattern=dots,pattern color=black] coordinates{(1,98.31) (2,96.74) (3,95.83) (4,95.58) (5,100)};
% De-En
\addplot+[color=black,fill=blue] coordinates{(1,98.37) (2,98.30) (3,97.75) (4,97.04) (5,100)};
% Fr-En
\addplot+[color=black,fill=blue,pattern color=blue,pattern=north east lines] coordinates{(1,98.05) (2,97.75) (3,97.26) (4,94.00) (5,100)};
\legend{Zh-En,De-En,Fr-En}
\end{axis}
\end{tikzpicture}
\caption{Statistics on the test sets, regarding documents without sentence-level coverage issue in document-level translation.}\label{fig:ana-cover}
\end{center}
\end{figure}

Following P-Transformer~\cite{p-transformer}, we employ a separator, $\#I$, to distinguish between the $I$-th and $(I+1)$-th sentences in both the source and target sides. 
By doing so, we can retrieve the sentence-level translation from the generated document using these separators. 
We consider a document to have no sentence-level coverage issues if we can perfectly recover sentence-level translation for all its source sentences.

From Figure~\ref{fig:ana-cover}, we find that 94\% to 98\% of documents are properly translated without sentence-level coverage issue.
For those few documents having sentence-level coverage issues, further analysis reveals that most of them miss one or two sentences at the end of translation. 
This shows that our proposed mixed-instruction can ensure that the sentences of the input document are translated accurately.

\section{Related Work}

We discuss related work from three perspectives: the document-level NMT models, the translation-oriented LLMs, and the position bias in LLMs.

\subsection{The Document-level NMT Models}

Conventional document-level NMT has achieved rich results.
These studies could be further roughly categorized into two groups. Studies of the first group extending the source-side context from a sentence to the local context of few sentences~\cite{tiedemann:2017,bawden_etal_naacl_2018,zhang-2018-doc,voita_etal_acl_2018,maruf-2018-doc,maruf:2019-doc,tan-etal-2021-coupling,lyu_et_al_emnlp:21,xu-etal-emnlp-2021-document-graph}. 
However, a notable disadvantage of these document-to-sentence models is that they struggle to effectively utilize the target-side context, which can be valuable for document-level translation~\cite{p-transformer}.
Studies of the second group have focused on document-to-document (Doc2Doc) translation. The initial exploration of Doc2Doc NMT has involved concatenating multiple sentences into a single unit~\cite{tiedemann:2017,agrawal:2018,ma:2020-simple,zhang-etal-2020-long}.
Recent studies have successfully trained vanilla Transformer models for Doc2Doc translation by leveraging either large augmented datasets~\cite{junczys:2019,lupo-2022-concate,sun-etal-2022-rethinking} or pre-trained models~\cite{liu-2020-mbart}. Additionally, some latest work has achieved the truly Doc2Doc translation by enhancing the attention mechanism in Transformer model~\cite{bao-etal-2021-g,p-transformer}.

\subsection{The Translation Oriented LLMs}

Research on translation-oriented LLMs can be broadly categorized into two categories. The first category focuses on empirical analysis, using LLMs as an interface~\cite{karpinska-iyyer-2023-large}. For instance, studies have evaluated the performance of ChatGPT, GPT3.5, GPT4, and text-davinci-002 in eighteen translation directions, including both high and low-resource languages~\cite{hendy_gpt-mt-2023}. Additionally, the work of~\cite{zhu2023multilingual} assessed the performance of popular open-source LLMs (XGLM, BLOOMZ, OPT) on 102 languages and compared them with strong supervised baselines.
A second approach to translation-oriented LLMs is fine-tuning smaller models (e.g., 7B model) for translation tasks. The work of~\cite{zeng2023tim} proposed using examples in comparison to teach LLMs to learn translation. The work of~\cite{chen2023improving} proposed improving the translation faithfulness of LLMs via augmenting instructions. Additionally, using large-scale sentence-level translation data to continue training LLMs~\cite{yang-etal-2023-BigTranslate,bayling}, which improved their translation performance. 
Different from these previous works, we propose to address the document-level translation problem by fine-tuning LLMs on translation mixed-instructions.

\subsection{The Position Bias in LLMs}

The causal language model used in many LLMs has a limitation on the use of long context information~\cite{liu2023lost}. 
For example, in multi-document question answering~\cite{liu2023lost} and text summarization~\cite{ravaut2023context}, it is difficult for LLMs to use the context information in the middle of documents. This position bias issue leads the LLMs to focus on initial or final parts, resulting a in U-shape performance pattern~\cite{ravaut2023context}. In machine translation, the work of~\cite{chen2023improving} argues that the position bias of LLMs increases the risk of instruction forgetting during decoding, resulting in hallucinations or unfaithfulness in translation.
We observed that the existing translation-oriented LLMs tend to struggle with effectively translating sentences in the middle and tail of a document. To address this position bias issue, we proposed a translation mixed-instructions, resulting in consistent translation performance from sentence level to as many as 2048 tokens documents.

\section{Conclusions}

In this paper, we have examined the primary reasons for the failure of document-level translation in the lightweight LLMs trained on sentence-level translation instructions. 
We find that most existing translation-oriented LLMs adopt sentence-level translation instruction during fine-tuning, resulting in insufficient document-level translation ability.
To address this issue, we propose a simple and effective fine-tuning approach using translation mixed-instructions including sentence-level and document-level instructions of varying lengths.
Our experimental results on several datasets show that our proposed translation mixed-instructions significantly outperform the strong baseline, achieving robust translation performance across both sentence-level and document-level translations.

Although our approach has achieved good document-level translation performance, its translation performance may slightly decrease when dealing with long documents (e.g., up to 2048 tokens).
In the future, we plan to verify our proposed approach on a broader range of open-source LLMs and investigate the document-level translation performance of LLMs on even longer documents beyond 2048 tokens.

\bibliographystyle{named}
%\bibliography{nmt}

\end{document}